\pdfoutput=1
\documentclass[11pt]{article}

\usepackage[table]{xcolor}
\usepackage{acl}

\usepackage{times}
\usepackage{latexsym}
\usepackage{booktabs} 
\usepackage{amsmath,bm}
\usepackage{enumitem}
\usepackage{multirow} 
\usepackage{amsfonts}
\usepackage{listings}
\usepackage{amsmath}
\usepackage[T1]{fontenc}
\usepackage[utf8]{inputenc}

\usepackage{microtype}

\usepackage{inconsolata}

\usepackage{graphicx}

\title{Instructions for *ACL Proceedings}

\author{Yijing Zhang, Dyah Adila, Changho Shin, Frederic Sala \\
        University of Wisconsin - Madison \\
        \{yzhang2637, adila, cshin23, fredsala\} @wisc.edu}

\newcommand{\SYSNAME}{\textsc{Chameleon}}
\newcommand{\norm}[1]{\left\lVert#1\right\rVert}

\title{Personalize Your LLM: Fake it then Align it}

\begin{document}
\maketitle
\begin{abstract}
Personalizing large language models (LLMs) is essential for delivering tailored interactions that improve user experience. Many existing personalization methods require fine-tuning LLMs for each user, rendering them prohibitively expensive for widespread adoption. Although retrieval-based approaches offer a more compute-efficient alternative, they still depend on large, high-quality datasets that are not consistently available for all users. To address this challenge, we propose \textbf{\SYSNAME}, a scalable and efficient personalization approach that uses (1) self-generated personal preference data and (2) representation editing to enable quick and cost-effective personalization.  Our experiments on various tasks, including those from the LaMP personalization benchmark, show that $\SYSNAME$ efficiently adapts models to personal preferences, improving instruction-tuned models and outperforms two personalization baselines by an average of 40\% across two model architectures. 

% using only a small subset of ground-truth personal data, offering a scalable solution while minimizing privacy risks.
\end{abstract}

\section{Introduction}

Large language models (LLMs) have transformed natural language processing (NLP), achieving excellent performance across a wide range of tasks. Their use has already expanded into diverse domains and user bases \cite{gururangan2020dontstoppretrainingadapt, shi2024medadapter, xu2024knowledge, xu2024ram}. This has motivated the need for personalization, i.e. tailoring these models to individual user preferences and specific contexts \cite{kirk2023personalisation}.

Current personalization methods are often impractical for large-scale deployment. Fine-tuning approaches \cite{li2024personalizedlanguagemodelingpersonalized, tan2024democratizinglargelanguagemodels, clarke2024peftuparameterefficientfinetuninguser} are resource-intensive, making it prohibitively expensive to customize models for each individual user. In contrast, retrieval-based methods \cite{salemi-etal-2024-lamp, di2023retrieval, fan2024survey} offer greater computational efficiency but suffer from a significant drawback: they rely on large high-quality datasets that are not consistently available for all users. These limitations impede the effective scaling of personalization, especially given the diverse and rapidly evolving nature of user preferences.

To achieve scalable personalization, we argue that two essential conditions must be met: (1) data efficiency, which enables effective personalization with minimal user interaction, and (2) compute efficiency, allowing for deployment across a large user base. We propose $\SYSNAME$, a new approach that fulfills both requirements by using synthetic, self-generated data to capture user preferences and uses representation editing to tailor its behavior to each user's unique preferences \cite{adila2024freeselfalignmentpossible}.

For each user, we begin with a small amount of historical data—sometimes as little as a single sample. Using this data, we prompt the LLM to generate two characteristic descriptions: one that reflects the user's personal preferences based on their history and another that represents a contrasting or non-personalized profile (e.g., "funny" versus "formal"). From these descriptions, we create synthetic user preference data. We then identify two distinct embedding spaces—personalized and non-personalized—derived from the synthetic preference data. Finally, we edit the LLM's embeddings to enhance the influence of the personalized subspace while diminishing the influence of the non-personalized subspace. 

With this data- and compute-efficient approach, we improve instruction-tuned models and two LLM personalization baselines by an average of $40 \%$ in the LaMP personalization benchmark \cite{salemi-etal-2024-lamp}. In summary, our contributions are:
\begin{enumerate}
    \item We introduce $\SYSNAME$, an LLM personalization framework that leverages self-generated user preference data and embedding editing techniques, providing \textbf{scalable, user-tailored personalization that is nearly cost-free}.

    \item On extensive evaluation using the LaMP benchmark \cite{salemi-etal-2024-lamp}, we show that $\SYSNAME$ improves upon instruction-tuned models and two LLM personalization benchmarks by an average of 40\% on two model architectures.

    \item $\SYSNAME$ can effectively personalize for new, unseen users without user history by leveraging profiles from other users with similar characteristics and preferences.
\end{enumerate}

% By leveraging self-generated preference data, our technique offers a \textbf{low-resource, almost free, scalable, user-centered approach} to LLM personalization, enabling more adaptive language models that can better serve diverse user needs.

\section{Related Work}
Our work seeks to address the personalization problem for LLMs using representation editing as an efficient technique to align models with user preferences. We give a brief overview of related areas.

\paragraph{Personalized LLMs.} 
Unlike general LLMs that produce uniform responses for all users, personalized LLMs adapt to the specific linguistic and communication preferences of individual users \cite{clarke2024peftuparameterefficientfinetuninguser}. Fine-tuning is a common method for achieving this, by training models on user-specific or task-specific data to personalize their behavior \cite{woźniak2024personalizedlargelanguagemodels}. Approaches like P-RLHF \cite{li2024personalizedlanguagemodelingpersonalized}, Persona-Plug \cite{liu2024llmspersonaplug}, and ALOE \cite{wu2024aligning} exemplify this strategy. However, fine-tuning is resource-intensive, making it impractical to personalize models for individual users at scale. Parameter-efficient fine-tuning (PEFT) \cite{tan2024democratizinglargelanguagemodels} reduces the computational burden but still requires large amounts of user data, which is often scarce and difficult to obtain in user personalization task \cite{zollo2024personalllm}.

Retrieval-based methods personalize model outputs by incorporating user-specific information retrieved at inference time \cite{dai2023uncovering,kang2023llms,liu2023chatgpt,wang2023learning,zhiyuli2023bookgpt,salemi-etal-2024-lamp}. While these methods avoid the need for tuning, they struggle with LLMs' limited context lengths, especially when dealing with long user histories. Although long-context models \cite{dubey2024llama, reid2024gemini, liu2024lost} allow for processing larger user histories, this incurs a high cost as many models are charged per token. Attempts to address this issue by summarizing retrieved information have been made \cite{richardson2023integratingsummarizationretrievalenhanced, liu2024once}. However, these approaches are vulnerable to distractions from irrelevant information \cite{shi2023large}, particularly when user behavior or preferences shift \cite{carroll2405ai, franklin2022recognising}.

The closest work to ours is LLM-REC \cite{lyu-etal-2024-llm}, a prompt-based approach that personalizes LLMs using summaries of selected top user history data. Our method takes this a step further by generating self-preference data, identifying embedding spaces that capture personalized versus non-personalized preferences, and performing personalization through representation editing. This enables a more data- and compute-efficient personalization process, making it possible to adapt models at scale to evolving user preferences quickly. Our approach represents a significant step toward scalable, real-time personalization that caters to dynamic user preference data.

% All approaches gain success in personalizing LLMs, yet all approaches still acquires large-scale user data, which performs limitations in performing efficient, rapid, group-scaled personalization.

% Under the rapidly emerging appication for LLMs and the changing user preferences from time to time, we foresee the need for rapid, large-scale personalization. Our work shows efficient model personalization in aligning with group-scaled users preferences.

% \paragraph{Self Data Generation} Obtaining high-quality human-annotated data can be significantly difficult. The problem can deteriorate in personalization scenario that annotated user preference data can be very limited. What's more, for personalization tasks, there is also raising concerns on privacy leaks in model parameters \cite{luo2024privacyllmbasedrecommendationrecent}. Synthetic datasets are then posed as a replacement solution to the challenge, and has shown deep potentials to further deploying into training \cite{voetman2023bigdatamythusing}. Approaches as \cite{wang2022self, sun2024principle} attempted to generate high-quality synthetic datasets using manually designed seed prompt as input for pretrained LLMs, and subsequently deploy the synthetic dataset for fine-tuning or training reward models. Yet no similar approaches touching the problem of personalizing LLMs, our work takes a novel approach through self-alignment based on self-generated user preference data and seeks solution for insufficient data and privacy concerns.

\paragraph{Representation Editing for Personalization.}
Representation editing has become an important technique for model alignment, involving the direct manipulation of a model’s latent representations to improve its performance and align it with desired attributes \cite{wang2024improvingtextembeddingslarge, kong2024aligninglargelanguagemodels}. For example, \citet{han2024wordembeddingssteerslanguage} demonstrated that steering LLM text embeddings can guide model output \emph{styles}. Similarly, \cite{li2024inference, han2023lm} show that adjusting embeddings during inference can enhance specific attributes, such as honesty or truthfulness, in the generated outputs.  \citet{liang2024controllabletextgenerationlarge} found that representation editing can control aspects of text generation, such as safety, sentiment, thematic consistency, and \emph{linguistic style}. These findings highlight the potential of using representation editing to guide models for personalization tasks. For visual generation models like Stable Diffusion, embedding-based personalization has long been recognized as an established technique \cite{han2023highly, arar2024palppromptalignedpersonalization, 10.1145/3618322, yang2024promptsoftboxpromptfreetextembeddingcontrol}.

\begin{figure*}[ht!]
    \centering
    \includegraphics[scale=0.27]{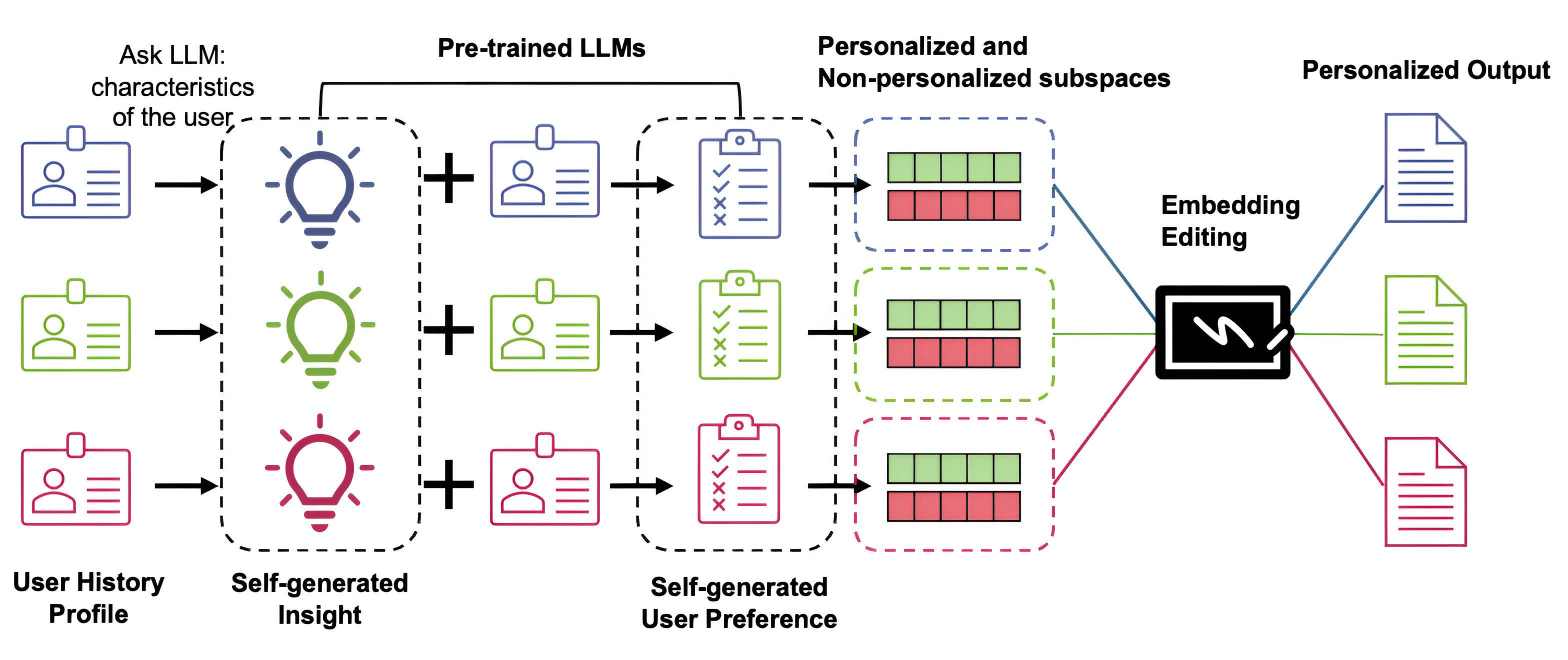}
    \caption{$\SYSNAME$ identifies two separate subspaces, one personalized and one non-personalized, from self-generated user characteristic insights. Based on these subspaces, we modify the LLM embeddings during inference.}
    \label{fig:flow}
\end{figure*}

Despite the growing interest in representation editing, little research has explored its application for personalizing LLMs, as proposed in our work. The most closely related study is \citet{adila2024freeselfalignmentpossible}, where the authors use embedding editing for general, rather than personalized, alignment to broad human preferences, relying on self-generated synthetic data. Our approach advances this notion by introducing a tailored mechanism that generates personalized synthetic data for each user and adapts embedding editing techniques for both individual and group-based personalization.

\begin{figure*}[t!]
    \centering
    % \hspace{-0.5cm}
    \includegraphics[scale=0.19]{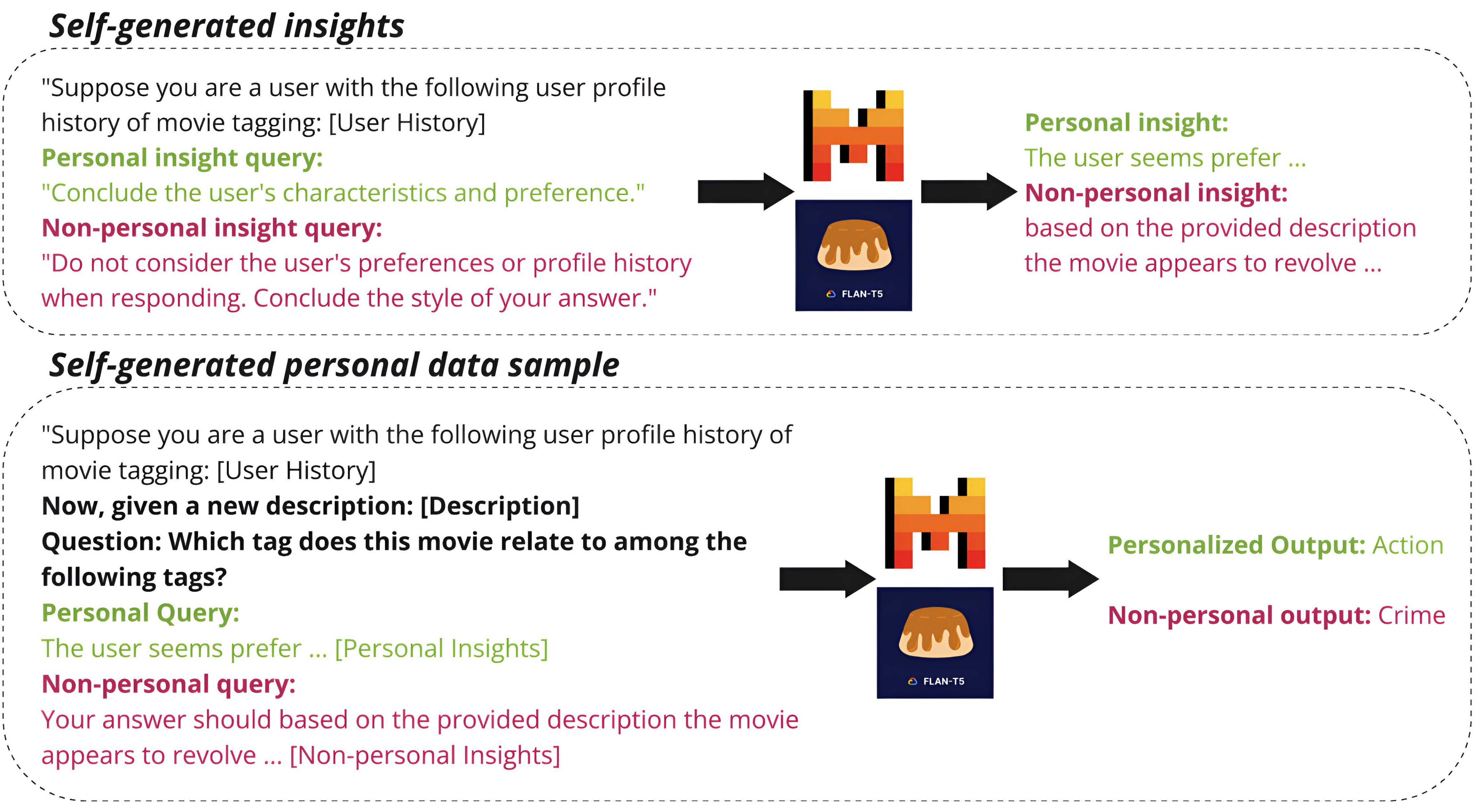}
    \caption{Self-generated user preference data: we use the generated conclusion of user characteristics to guide the personal answer generation.}
    \label{fig:generate_pref_data}
\end{figure*}

\section{\SYSNAME: Personalization through Representation Editing}

% Personalized large language models (LLMs) aim to meet each user's unique needs and preferences by adapting responses based on their past behaviors \cite{tan2024democratizinglargelanguagemodels, salemi-etal-2024-lamp}. More concretely, following previous studies, personalizing LLMs can be defined as: for each user $u$, given user history data $\mathcal{H}_u = \{h_u^i\}$ and an input query $q_u$, generate a personalized output $\hat{y}_u$ (i.e. $\hat{y}_u = f(\mathcal{H}_u; q_u)$) \cite{salemi-etal-2024-lamp, liu2024llmspersonaplug}. For a scaled universal model for multiple users, correspondingly, we should find the minimum sum of personalization loss (deviation) $L$ between the generated output $\hat{y}_u$ and the reference output $y_u$ given $\mathcal{H}_u$ and $q_u$ for each user $u$:
% \begin{equation}
%     \min \sum_u L(\mathcal{H}_u; q_u; y_u)
% \end{equation}
% \cite{zhuang2024hydramodelfactorizationframework}. 
% Current research mostly focus on calculating and minimizing $L(\mathcal{H}_u; q_u; y_u)$ to finetune models separately for each user $u$, which can be costly \cite{salemi2024comparingretrievalaugmentationparameterefficientfinetuning} 
%and create privacy conerns \cite{luo2024privacyllmbasedrecommendationrecent}

% Thus, 
We present \SYSNAME, an almost \textbf{cost-free} alignment personalization framework with representation editing using self-generated synthetic user preference data. Figure \ref{fig:flow} illustrates our technique. We achieve personalization with two stages: (1) self-generating user preference data (Section \ref{sec:generated_pref_data}), and (2) representation editing using the self-generated data (Section \ref{sec:representation_editing}). 
Additionally, we extend $\SYSNAME$  to support \textbf{scalable user groups}, enabling efficient alignment at a group level (Section \ref{sec:group_scale_alignment}).
% for \textbf{scalable users group}

% \begin{figure*}[t!]
%     \centering
%     \includegraphics[width=1\textwidth]{figs/generate_data2.png}
%     \caption{Generating preference pairs. First, we prompt pretrained models to conclude user's \textit{insight} on the personal characteristics based on their personal history profile (top). Then, we use the generated conclusion of user characteristics to guide the personal answer generation (bottom).}
%     \label{fig:generate_pref_data}
% \end{figure*}

\subsection{Self-generated Preference Data}
\label{sec:generated_pref_data}

Our method for generating self-preference data uses generic, non-personalized LLMs to identify user-specific characteristics and preferences from the available user history. Using these identified characteristics, we prompt the model to generate tailored responses for each user. This process consists of three key steps: (1) selecting relevant user history, (2) generating insights from the selected history, and (3) producing synthetic user preference data guided by these insights.

\paragraph{User History Selection.} 
User's historical behavior usually contains important information regarding their characteristics, linguistic patterns, and preferred interactions. However, not all historical behaviors serve as reliable indicators of user preferences. Adapting the model using redundant and generic user behavior may not result in high-quality personalized LLMs. Selecting and filtering for representative user historical behavior is thus important. Although recent studies showed success in using retrieval-based re-rankers \cite{zhuang2024hydramodelfactorizationframework} and encoder-based user history selection \cite{liu2024llmspersonaplug}, they can struggle when user preferences shift rapidly or when there’s limited historical data. To address this, we focus on a more lightweight and adaptable approach to user history selection.

Since our approach relies on embedding editing to adapt the model, we need to identify user-representative historical data. The first step is to define what makes this data "representative." We leverage sentence embeddings for their strong ability to capture both the meaning and context of text \cite{reimers2019sentencebertsentenceembeddingsusing}. Our goal is to find the most informative and relevant embedding pieces that reflect key user preferences. A lightweight approach to find such data is to perform principal component analysis (PCA) on the embeddings \cite{Gewers_2021}. Specifically, for each user $u$, given a set of user history $\mathcal{H}_u = \{ {h_u^i} \}$ where each $h_u^i$ represents an individual user history sample with index $i$, we have
\begin{equation}
   e_u^i = \textbf{SentenceEmbedder}(h_u^i) .
\end{equation}

Then, we have that $\bm{W}_u$ are the top $k$ principal components of $\bm{E_u} = [e_u^1, e_u^2, \dots,  e_u^N]^\top$ and the projection of each embedding is $z_u^i = e_u^i \bm{W}_u$. We next find the top $k$ history data embeddings: 
\begin{equation}
    \bm{E}_u^{k} = \underset{i \in [1,\ldots, N]}{\arg\text{top-}k} \norm{z_u^i},
\end{equation}
and get top $k$ history data $\bm{H}_u^k = \{ h_u^i: i \in \bm{E}^k_u \}$.

\paragraph{Insight Generation.}

We query an instruction-tuned general-purpose LM to analyze and infer characteristics specific to individual users. For each user $u$, given the selected set of user history $\bm{H}_u^k$ from the previous step, we query the LM (denoted as $\omega$) and generate two distinct styles of responses: one as a personalized agent (\( C^{P} \)) and the other as a non-personalized/neutral agent ($C^{N}$). The personalized agent ($C^{P}$) draws on the user's historical data $\bm{H}_u^k$, concluding insights about the user's preferences, behaviors, and style. The neutral agent ($C^{N}$) is asked to give characteristics of impersonal and general responses. It represents the standard behavior of the model when user personalization is absent. Then, for each user $u$, we have an personalized-neutral insights pair $(c_u^P, c_u^N)$.

% choose user history based on embedding
% we intend to find two "opposite" subspace on personalizatin task
% prompting templating for guiding the generation for insight for different subspaces

% \begin{figure}[t!]
%     \centering
%     \includegraphics[width=1\textwidth]{figs/self_generated.png}
%     \caption{Self-generating user preference data: we use the generated conclusion of user characteristics to guide the personal answer generation.}
%     \label{fig:self_generated}
% \end{figure}

\paragraph{Generating Synthetic User Preference Data}

Once the insights are generated, we use the insight pairs as prompt guidance to generate synthetic user preference data. For each user $u$ and each user query $q_u$, given the pre-selected history set $\mathcal{H}_u$ and insight pair $(c_u^{i, P}, c_u^{i, N})$, we have our general-purpose LM ($\omega$) separately generate personalized and neutral preference outputs $(\hat{y}_u^{i, P}, \hat{y}_u^{i, N})$ to query $q_u^i$ conditioned on $(c_u^{i,P}, c_u^{i,N})$ respectively. We then concatenate the outputs $(\hat{y}_u^{i,P}, \hat{y}_u^{i,N})$ with user history $\mathcal{H}_u$ and obtain the self-generated preference pair $(p_u^{i,P}, p_u^{i,N})$ for each user query $q_u^i$. By applying this procedure to all user queries, we obtain self-generated preference data pairs $(P_u^{P}, P_u^{N})$.

Note that we do not apply any prompt tuning; rather, we use a predefined set of prompt templates and a frozen LLM for all generations. Figure \ref{fig:generate_pref_data} illustrates the full process, with prompting details in Appendix \ref{appendix:prompt}.

% For large-scale multi-user alignment at once, we continue this process for all users in the group, and combine all self-generated preference data pair and get $(P^P, P^N) = \{(P^P_u, P_u^N), u \in U\}$. We will then use this set for representaion editing.

\subsection{Representation Editing}
\label{sec:representation_editing}

Next, using the self-generated user preference data, we align the model with users' preferences with a technique inspired by \textsc{AlignEZ} \cite{adila2024freeselfalignmentpossible}. We first identify two subspaces in the model's embedding space (denoted as vector $\theta \in \mathbb{R}^d$ in LM $\omega$'s latent space) that correspond with the users' preferences. We use singular value decomposition (SVD) on the preference data embeddings to capture directions of the personalized embeddings $\theta^P_{l, u}$. Next, we employ CCS-based identification \cite{burnsccs} to find the hyperplane that best separates the non-personalized embeddings from the personalized ones and denote the directions of the hyperplane as $\theta^N_{l,u}$. A detailed explanation is provided in Appendix \ref{appsubsec:rep_edit}.

With the personalized and non-personalized subspaces $\theta^P$ and $\theta^N$, we perform embedding editing on the MLP outputs of the most impactful decoder layers (i.e. layers that have lowest average CSS loss) during the inference phase to adapt the LLM to users' preferences. More concretely, given $x_l$, the output of the MLP of layer $l \in L$, where $L$ is the set of layers with lowest average CSS loss, we strengthen the personalized direction by
\[\hat{x}_{l,u} \leftarrow x_l + \frac{\langle x_l , \theta_{l, u}^P \rangle}{\langle \theta_{l,u}^P, \theta_{l,u}^P \rangle} \theta_{l,u}^P\]
and remove the non-personalized direction by
\[\hat{x}_{l,u} \leftarrow \hat{x}_{l,u} - \frac{\langle \hat{x}_{l,u} , \theta_{l,u}^N \rangle}{\langle \theta_{l,u}^N, \theta_{l,u}^N \rangle} \theta_{l,u}^N.\]
These edits are performed for each user query.

\subsection{Group-scale Personalization}
\label{sec:group_scale_alignment}

% When handling multiple users simultaneously, aligning the model individually for each user would be inefficient \cite{dai2024mpcodermultiuserpersonalizedcode}. To address this challenge, instead of performing alignment for each individual user, we combine all users with history data as a group and perform a one-time collective group-scale alignment. More specifically, we combine the synthetic self-preference data of all users into one set $(P^P, P^N) = \{(P^P_u, P_u^N), u \in U\}$, which will used for finding direction vectors in representation editing.

% This method enables efficient multi-user alignment by processing all requests simultaneously, allowing quick personalization. In Section \ref{sec:number}, we empirically show that group personalization improves performance compared to the single-user setting. Additionally, group-scale personalization also allows us to use available data from other users for users that have no history data yet, allowing personalization for even unseen users (Experiment \ref{sec:experimental_results}).

Individually aligning the model for multiple users is inefficient when scaling to a large user base \cite{dai2024mpcodermultiuserpersonalizedcode}. To overcome this, we extend \SYSNAME \ to  group-scale alignment. Instead of aligning for each user separately, we combine the history data of all users into a single group and perform collective alignment. Specifically, we aggregate the synthetic self-preference data for all users into one set, $(P^P, P^N) = \{(p_u^{i,P}, p_u^{i,N}) \in (P^P_u, P^N_u)| u \in U\}$, where $U$ is the set of users in the group. $(P^P, P^N)$ is then used to find direction vectors for representation editing.

This approach enables efficient personalization by processing all users simultaneously, leading to faster alignment. In Section \ref{sec:number}, we show that group-scale personalization outperforms the single-user setting. Furthermore, this method allows us to leverage data from other users for those with no available history, enabling personalization for new or unseen users (see Experiment \ref{sec:experimental_results}).

\section{Experiments}

\begin{table*}[t!]
   \centering
   \scalebox{0.89}{
    \begin{tabular}{c | c || c | c | c | c || c | c | c | c}
    \toprule
                   \multicolumn{2}{c}{\textbf{Models $\rightarrow$ }}      & \multicolumn{4}{c}{Mistral Instruct}                                 & \multicolumn{4}{c}{Flan T5 XXL} \\
            \hline \hline
    \multirow{2}{*}{Dataset} & \multirow{2}{*}{Metric}   & Instruct & LLM      & \multirow{2}{*}{ALOE}       &  \multirow{2}{*}{\textbf{$\SYSNAME$}}            & Instruct    & LLM & \multirow{2}{*}{ALOE} &  \multirow{2}{*}{\textbf{$\SYSNAME$}}        \\
     &   & Model & -REC      &      &   & Model   & -REC & &        \\
    \hline \hline
    & Acc. $\uparrow$ & 0.198 & 0.262                                           &  0.307 & \cellcolor{blue!20}{\textbf{0.396}} &  0.238  &  0.214  & 0.333 & \cellcolor{blue!20}{\textbf{0.420}} \\
    \multirow{-2}{*}{LaMP2}      & F-1 $\uparrow$      & 0.236 & 0.309         & 0.220 &  \cellcolor{blue!20}{\textbf{0.349}}  & 0.171    &  0.146  & 0.255 & \cellcolor{blue!20}{\textbf{0.311}}  \\ 
    & MAE $\downarrow$ & 0.497 & 0.484                                           & 0.423 &   \cellcolor{green!20}{\textbf{0.407}} & 0.456  &  0.798  & 0.427 & \cellcolor{green!20}{\textbf{0.400}} \\
    \multirow{-2}{*}{LaMP3}      & RMSE $\downarrow$      & 0.944 &   0.976     & 0.888  &   \cellcolor{green!20}{\textbf{0.815}} & 0.818  &  1.439  & 0.786 & \cellcolor{green!20}{\textbf{0.714}}  \\ 
    & R-1 $\uparrow$ & 0.354 & 0.183 & 0.362 & \cellcolor{blue!20}{\textbf{0.381}}   &  0.333  & 0.225 & 0.376 & \cellcolor{blue!20}{\textbf{0.429}}    \\
    \multirow{-2}{*}{LaMP7}      & R-L $\uparrow$      & 0.295 & 0.144 & 0.313 & \cellcolor{blue!20}{\textbf{0.334}} & 0.292 & 0.196 & 0.331 & \cellcolor{blue!20}{\textbf{0.385}}    \\ 
    \bottomrule
    \end{tabular}
    }
    \caption{$\SYSNAME$ outperforms all baselines in personalization for users with history. Best performance is highlighted in \textbf{bold}. Metrics where higher values indicate better performance are shaded in \colorbox{blue!20}{blue cells}, while metrics where lower values are preferable are marked with \colorbox{green!20}{green cells}.
    }
    \label{tab:results}
\end{table*}

\begin{table*}[t!]
   \centering
   \scalebox{0.89}{
    \begin{tabular}{c | c || c | c || c | c }
    \toprule
                   \multicolumn{2}{c}{\textbf{Models $\rightarrow$ }}      & \multicolumn{2}{c}{Mistral Instruct}                                 & \multicolumn{2}{c}{Flan T5 XXL} \\
            \hline \hline
    Dataset & Metric & ALOE & \textbf{$\SYSNAME$} & ALOE &  \textbf{$\SYSNAME$}      \\
    \hline \hline
    & Acc. $\uparrow$ & 0.227 & \cellcolor{blue!20}{\textbf{0.363}}    & 0.109 & \cellcolor{blue!20}{\textbf{0.390}}    \\
    \multirow{-2}{*}{LaMP2}      & F-1 $\uparrow$ & 0.177 &  \cellcolor{blue!20}{\textbf{0.338}}  & 0.040 & \cellcolor{blue!20}{\textbf{0.304}}    \\ 
    & MAE $\downarrow$ & 0.522 &   \cellcolor{green!20}{\textbf{0.442}} & 0.544 & \cellcolor{green!20}{\textbf{0.413}}    \\
    \multirow{-2}{*}{LaMP3} & RMSE $\downarrow$ & 0.906 & \cellcolor{green!20}{\textbf{0.903}} & 1.030   &  \cellcolor{green!20}{\textbf{0.839}}    \\ 
    & R-1 $\uparrow$ & 0.185 & \cellcolor{blue!20}{\textbf{0.377}}   & 0.251 & \cellcolor{blue!20}{\textbf{0.420}}    \\
    \multirow{-2}{*}{LaMP7}  & R-L $\uparrow$ & 0.155 & \cellcolor{blue!20}{\textbf{0.331}} & 0.206 & \cellcolor{blue!20}{\textbf{0.373}}    \\ 
    \bottomrule
    \end{tabular}

    }
     \caption{$\SYSNAME$ performance compared ALOE on new unseen users.}
    \label{tab:results_unseen}
\end{table*}

We begin by detailing our experimental setup in Section \ref{sec:exp_setup}, followed by experiments to validate the following key claims about $\SYSNAME$:
\begin{itemize}[leftmargin=*]
\itemsep0em 
    \item Aligns LLMs to user-specific preferences (Section \ref{sec:experimental_results}),
    \item Generalizes to unseen users (Section \ref{sec:unseen_users}),
    \item Group-scale personalization improves performance (Section \ref{sec:number}),
    \item Outperforms compute extensive methods like DPO in time-constrained scenarios (Section \ref{sec:quick}).
    % \item Ablation (Section \ref{sec:ablations}).
\end{itemize}
In Section \ref{sec:ablations}, we perform ablation study to understand the effect of the number of user history data to $\SYSNAME$ performance.

\subsection{Experimental Setup}
\label{sec:exp_setup}

\paragraph{Datasets and Tasks.} We evaluate $\SYSNAME$ using the LaMP language model personalization benchmark \cite{salemi-etal-2024-lamp}. Our evaluation focuses on three specific personalization tasks: (1) Personalized Movie Tagging (LaMP 2), (2) Personalized Product Rating (LaMP 3), and (3) Personalized Tweet Paraphrasing (LaMP 7). We adhered to the user-based data split provided by the LaMP benchmark, using the default training and test splits. Additional details about the datasets and tasks can be found in Appendix \ref{appendix:dataset}.

\paragraph{Evaluation Metrics.}
We use the evaluation metrics established by the LaMP benchmark for each task. For Personalized Movie Tagging (LaMP 2), we measure Accuracy (Acc.) and F-1 Score (F-1). For Personalized Product Rating (LaMP 3), we assess performance using Mean Absolute Error (MAE) and Root Mean Squared Error (RMSE). For Personalized Tweet Paraphrasing (LaMP 7), we apply the ROUGE-1 (R-1) and ROUGE-L (R-L) metrics.

\paragraph{Baseline 1: Non-personalized Instruction-tuned Models.} We evaluate $\SYSNAME$ against two general purpose instruction-tuned models: Mistral-7B-v0.3-Instruct \cite{jiang2023mistral7b} and Flan-T5 XXL \cite{chung2022scalinginstructionfinetunedlanguagemodels}. Both models are assessed using the same set of user queries as $\SYSNAME$, following the same prompt format and using the same pre-selected user history profile—excluding any insights. Additional prompt details can be found in Appendix \ref{appendix:prompt}.

\paragraph{Baseline 2: Personalization Methods.} 
We also compare $\SYSNAME$ against two personalization techniques, namely LLM-REC \cite{lyu-etal-2024-llm}, a prompting-engineering personalization method, and ALOE \cite{wu2024aligning}, a supervised Fine-tuning (SFT) personalization method.
% via structuring tree-shape personalization interaction profile, over the LaMP benchmark. 

% \paragraph{Group Personalization Implementation.} We implement group personalization setting (Section \ref{sec:group_scale_alignment} as follows: First, we randomly pick 100 users from the training split of the LaMP benchmark dataset and use the PCA-base history selection (Section \ref{sec:generated_pref_data}) to select at most 10 user histories from the user profile. For each of the selected 100 users, we separately query for personalized and neutral insight pairs and the self-generated preference data. We discard self-generated data pieces where personalized and non-personalized output is the same. We then concatenate the self-generated user preference data, perform self-alignment, and test the performance of the personalized aligned model on unseeen user queries (the validation split of LaMP dataset \cite{salemi-etal-2024-lamp}). We perform the random 100 user selection repeatedly and compute the average performance.

\paragraph{Group Personalization Setup.} To implement group-scale personalization (Section \ref{sec:group_scale_alignment}), we randomly select 100 users from the training split of the LaMP benchmark. Using PCA-based history selection (Section \ref{sec:generated_pref_data}), we choose up to 10 user history entries per profile. For each user, we generate personalized and neutral insight pairs along with self-generated preference data. Any data where the personalized and non-personalized outputs are identical is discarded. We then combine the self-generated preference data for all users, perform group-scale alignment, and evaluate the personalized model on unseen user queries from the LaMP test split (Section \ref{sec:unseen_users}). This process is repeated for different random sets of 100 users, and we report the average performance.

% Instead of taking one user and performing individual model adaption for each user, our work focused on taking multiple users as a group at once and performing one single alignment for all users (Section \ref{sec:group_scale_alignment}). 
% Also, instead of feeding large-scale user history data to the model, our work only requires a small amount of personal data. 
\subsection{Aligns LLMs to user-specific preferences}
\label{sec:experimental_results}

\paragraph{Setup.}
We compare $\SYSNAME$ with the previously mentioned baselines. In the self-insight generation process, user history data is fed directly to the models using simple prompts (see Appendix \ref{appendix:prompt}), without access to human annotations.

\paragraph{Results.} As shown in Table \ref{tab:results}, $\SYSNAME$ consistently outperforms all baselines. Remarkably, these improvements are achieved with minimal user history data and without any training and fine-tuning, surpassing an SFT-based method (ALOE). \textbf{These results validate our claim that $\SYSNAME$ can effectively align LLMs to individual user preferences}.

\subsection{Generalizes to unseen users}
\label{sec:unseen_users}

\paragraph{Setup.} We also assess $\SYSNAME$'s ability to personalize for new, unseen users who have no prior history. In this evaluation, we run both $\SYSNAME$ and ALOE on the LaMP training split and evaluate their performance on test samples from users not included in the training data. This experimental setup is not applicable to instruct models and LLM-REC, as both of these methods use prompt-based personalization and do not differentiate between seen and unseen users.
% With new unseen users, we still provide user history data in required format for both approaches. 

% [\textcolor{red}{TODO: Jane explain what you did for unseen users on Chameleon and ALOE, also why this setup is not applicable for the other 2 baselines}]. 

\paragraph{Results.} Table \ref{tab:results_unseen} demonstrates that $\SYSNAME$ achieves strong personalization performance even with new, unseen users, \textbf{validating our claim that $\SYSNAME$ can effectively generalize to users without prior history}. In contrast, ALOE struggles in this scenario, suggesting that it may overfit to the characteristics of users in the training set.

\subsection{Group-scale personalization improves performance}
\label{sec:number}

\paragraph{Setup.} To assess the effectiveness of group-scale personalization compared to single-user personalization, we run $\SYSNAME$ on groups of varying sizes. We experiment with group sizes of $\{1, 20, 40, 60, 80, 100\}$ on both LaMP2 and LaMP3 tasks, while keeping the amount of generated insights and preference data per user constant.

\begin{figure}[t!]
    \centering
    \includegraphics[scale=0.63]{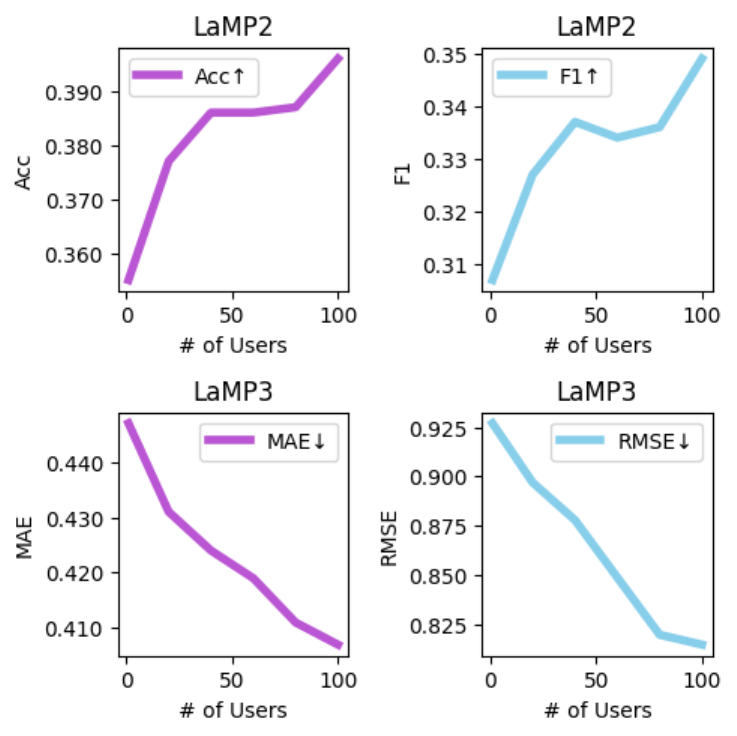}
    \caption{The change of performance when different number of users are given to $\SYSNAME$}
    \label{fig:user_number}
\end{figure}

\paragraph{Results.} Figure \ref{fig:user_number} reveals a clear trend: as the number of users in the group increases, personalization performance consistently improves. This is evident both when shifting from a single-user setup (left-most point, where number of users = 1) to group personalization, and as the group size grows. \textbf{These results support our claim that group personalization offers performance gain compared to single-user personalization}.

\subsection{Outperforms DPO in time-constrained scenario}
\label{sec:quick}

\begin{figure}[t!]
    \centering
    \includegraphics[scale=0.49]{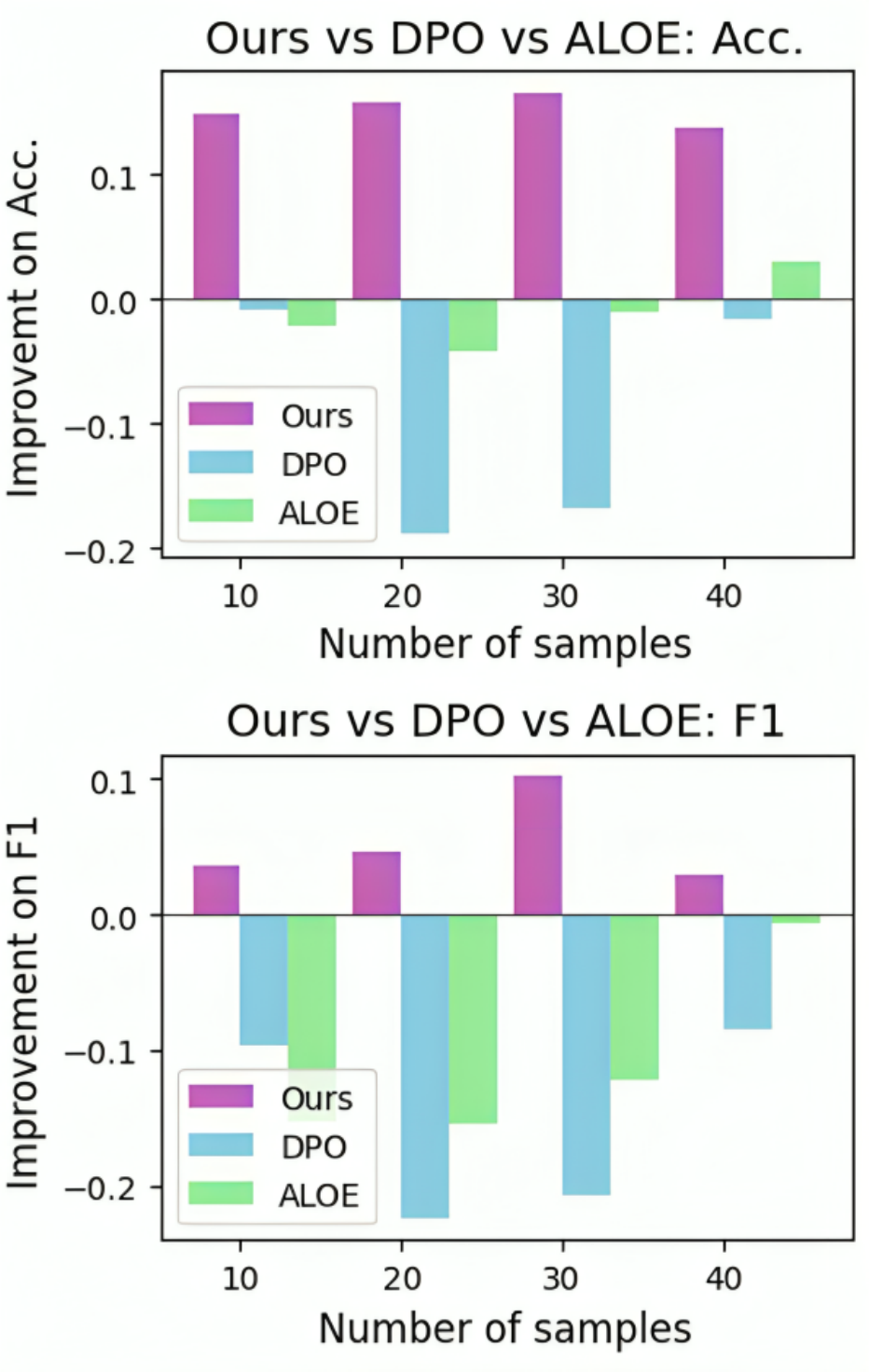}
    \caption{$\SYSNAME$ compared with DPO and ALOE in time-constrained scenarios. The columns denotes the improvement from the instruction-tuned model.}
    \label{fig:time}
\end{figure}

\paragraph{Setup.} We compare $\SYSNAME$ with DPO \cite{rafailov2024directpreferenceoptimizationlanguage} and ALOE \cite{wu2024aligning}, a tuning-based alignment and SFT-based personalization methods, in a time-constrained scenario where alignment must be performed quickly. In this setup, we fix the time allowed for all methods and get the number of samples for each method within that time. This setup reflects real-world situations where instant personalization is required for new users with little to no available data. Hyperparameter details for DPO and ALOE are provided in Appendix \ref{appendix:time}.

\paragraph{Results} As shown in Figure \ref{fig:time}, $\SYSNAME$ consistently delivers stable personalization gains in the time-constrained scenario, whereas both ALOE and DPO struggle with limited sample availability. This supports our claim that \textbf{$\SYSNAME$ is more suitable than resource-intensive approaches in time-sensitive scenarios}.

\subsection{Editing both personalized and non-personalized embedding improves performance.}
\label{sec:embedding}

\paragraph{Setup.} To examine the individual effects of personalized and non-personalized profile, we conducted an experiment on only editing personalized/non-personalized embedding space on the LaMP2 and LaMP3 tasks on Mistral instruct models.

\paragraph{Results.} We report the metric for each case in Table \ref{tab:results_embedding}. CHAMELEON rely on editing in personalized embedding space to give personalized outputs, and removing non-personalized embedding space follows previous studies that removing spurious or unwanted concept subspaces from embeddings boosts model accuracy on rare class predictions \cite{adila2024zeroshotrobustificationzeroshotmodels, chuang2023debiasingvisionlanguagemodelsbiased}.

\begin{table}
   \centering
   \scalebox{0.80}{
    \begin{tabular}{c | c || c | c | c}
    \toprule
                   \multicolumn{2}{c}{\textbf{Models $\rightarrow$ }}      & \multicolumn{2}{c}{Mistral Instruct}                               \\
            \hline \hline
     & & Only & Only Non- & \\ 
    \multirow{-2}{*}{Dataset}  & \multirow{-2}{*}{Metric}  & personalized & personalized & \multirow{-2}{*}{Both}  \\
    \hline \hline
    & Acc. $\uparrow$ & 0.356 & 0.346 & \cellcolor{blue!20}{\textbf{0.396}} \\
    \multirow{-2}{*}{LaMP2}      & F-1 $\uparrow$ & 0.276 & 0.268 &  \cellcolor{blue!20}{\textbf{0.349}}  \\ 
    & MAE $\downarrow$ & 0.484 & 0.494 &   \cellcolor{green!20}{\textbf{0.407}} \\
    \multirow{-2}{*}{LaMP3} & RMSE $\downarrow$ & 0.900 & 1.005 & \cellcolor{green!20}{\textbf{0.815}} \\ 
    \bottomrule
    \end{tabular}
    }
     \caption{Embeddings to edit effect to performance of $\SYSNAME$.}
    \label{tab:results_embedding}
\end{table}

\subsection{Ablations}
\label{sec:ablations}

\paragraph{Setup.} To examine the impact of the amount of user history data on performance, we run $\SYSNAME$ on the LaMP2 and LaMP3 task, varying the number of history per user as $\{5,10,15,20,25\}$, while keeping the number of users in the group constant.

\begin{figure}[t!]
    \centering
    
    \includegraphics[scale=0.39]{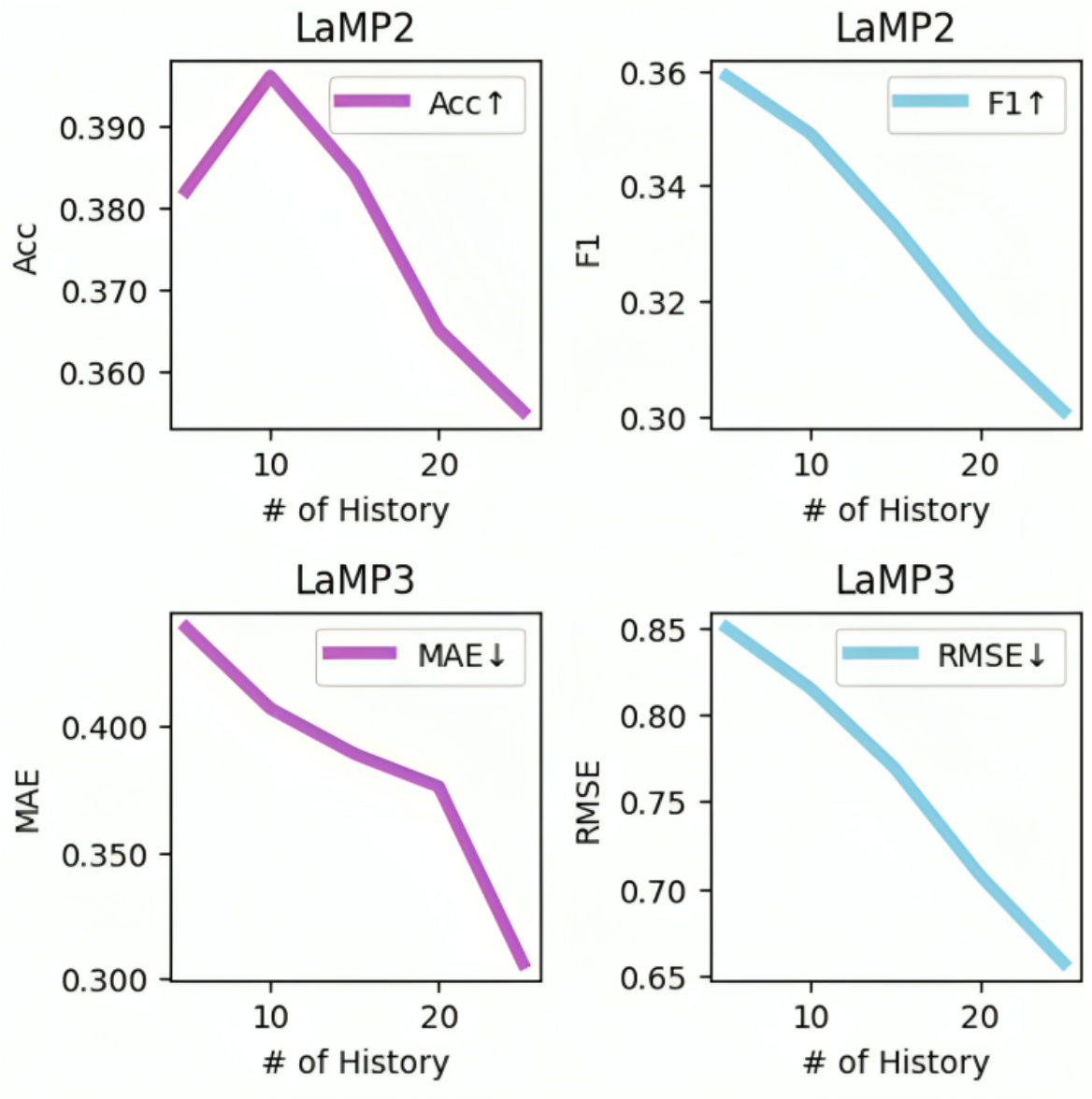}
    \caption{The change of performance when different number of history data per user are given to $\SYSNAME$}
    \label{fig:his_num}
\end{figure}

\paragraph{Results.} Figure \ref{fig:his_num} illustrates that when the amount of user history data is small, the performance improvement of $\SYSNAME$ is limited. This limitation likely arises from the difficulty in generating accurate personalization insights with insufficient data. Conversely, when the amount of history data is too big, the performance of $\SYSNAME$ declines. We hypothesize that this deterioration occurs because too many history profiles may introduce unrelated or outdated samples, hindering effective personalization. In most cases, only a small number of history data would lead to a good performance for $\SYSNAME$.

% likely due to accumulated noise to find personalization embedding for all users. We found an optimal history number at around 10.

\section{Discussion}

\paragraph{Limitations.} 
While $\SYSNAME$ successfully delivers scalable personalization with minimal costs, it has some limitations. A key challenge is its dependence on the quality of the self-generated preference data. Although aligning the model with this data yields promising results, the effectiveness of the personalization largely depends on how accurately and comprehensively user preferences are captured by the base LLM. Future research could focus on developing more refined metrics to capture personal characteristics better, ensuring more precise and reliable self-alignment.

One potential risk with $\SYSNAME$ is the possibility of malicious input in user history. Since $\SYSNAME$ relies on a limited amount of user history to generate self-preference data for alignment, harmful or biased history inputs could unintentionally lead the model to produce toxic or malicious responses. This highlights the need for strong safeguards, such as thorough filtering and ethical review processes, to prevent the model from aligning with or reinforcing negative behaviors while still delivering effective personalization.

\paragraph{Ethical Considerations.}

Privacy has long been a problem for LLM personalization, as personalizing LLMs usually require large-scale personal data and preferredly (human) labeled, which could lead to potential privacy leaks. Though personalization dataset, like LaMP benchmark dataset used in our experiments, is publicly accessible an does not raise privacy concerns, personal data collection and usage still presents significant challenge in personalizing LLMs. With our approach, we only acquire a very small portion of user historical data and resolve data labeling problem with self-generation technique. And since self-generated user preference data are fake synthetic data for performing alignment, it can possibly reduce the risk of privacy leaks.

\paragraph{Conclusion.}

We present $\SYSNAME$, a novel light-weight, scalable approach for personalizing LLMs without access to large-scale human-annotated personal data and individual fine-tuning. By leveraging the ability to conclude and capture user characteristics and preferences, $\SYSNAME$ adjusts the model embeddings during inference to generate outputs that are more aligned with user preferences. Our experiments show that $\SYSNAME$ significantly enhance the personalization ability of base language models using only a small portion of real user data, and it is able to adapt models with multiple user expectations within one single alignment process.

This work represents an initial step toward achieving cost-free, rapid, group-scale personalization that current personalization methods struggle to address.

\bibliography{bib}

\appendix

\section{Appendix}

\subsection{Glossary}
Table \ref{table:glossary} shows glossary of terms used in this paper.
\label{appendix:glossary}
\label{sec:gloss}
\begin{table*}[]
\centering
\begin{tabular}{l l}
\toprule
Symbol & Definition \\
\midrule
$y$ & Ground truth output \\
$\hat{y}$ & Model prediction \\
$\mathcal{H}_u$ & Set of user history for user $u$ \\
$h^i_u$ & i-th user history for user $h$ (i-th data data in $\mathcal{H}_u$) \\
$e^i_u$ & Sentence embedding of $h^i_u$ \\
$\bm{E}_u$ & Embedding matrix of user history for user $u$ \\
$\bm{H}_u^k$ & Top $k$ selected history data \\
$C^P$ & Personalized agent \\
$C^N$ & Non-personalized agent \\
$c^P_u$ & Personalized insights for user $u$ \\
$c^N_u$ & Non-personalized insights for user $u$ \\
$c^{i, P}_u$ & i-th personalized insight for user $u$ \\
$c^{i, N}_u$ & i-th non-personalized insight for user $u$ \\
$\hat{y}^{i,P}_u$ & Model prediction conditioned on $c^{i, P}_u$ \\
$\hat{y}^{i,N}_u$ & Model prediction conditioned on $c^{i, N}_u$ \\
$q_u^i$ & i-th query for user $u$ \\
$p_u^{i,P}$ & Personalized preference for user query $q^i_u$ \\
$p_u^{i,N}$ & Non-personalized preference for user query $q^i_u$ \\
$P^P_u$ & Set of personalized preferences for user $u$ \\
$P^N_u$ & Set of non-personalized preferences for user $u$ \\
$\theta^P$ & Personalized embedding direction \\
$\theta^N$ & Non-personalized embedding direction \\
$\theta^P_{l,u}$ & Personalized embedding direction for user $u$ at layer $l$\\
$\theta^N_{l,u}$ & Non-personalized embedding direction for user $u$ at layer $l$ \\
$x_l$ & Representation (embedding) at layer $l$ \\
$\hat{x}_{l,u}$ & Personalized representation for user $u$ at layer $l$ \\

\toprule
\end{tabular}
\caption{
	Glossary of variables and symbols used in this paper.
}
\label{table:glossary}
\end{table*}

\subsection{Dataset and Task Details}
\label{appendix:dataset}
The LaMP dataset is a publicly available dataset for personalizing LLMs. We only used LaMP dataset for the purpose of running the experiments.

The tasks of LaMP we experimented with are as follows:
\begin{enumerate}
    \item \textbf{LaMP 2: Personalized Movie Tagging.} Given a user profile of user history tagging along with the movie description, you are tasked to predict the movie tag given a new movie description.
    \item \textbf{LaMP 3: Personalized Product Rating.} Given a user profile of user history product rating along with the product reviews, you are tasked to predict the rating of a product given a new product review wrote by the user.
    \item \textbf{LaMP 7: Personalized Tweet Paraphrasing.} Given a user profile of user history tweets you are tasked to predict how the user may paraphrase a new given tweet.
\end{enumerate}

Details of LaMP dataset is presented in Table \ref{tab:lamp}. $[\textit{italic text}]$ presents actual data. 

\begin{table*}[ht]
\caption{LaMP Dataset Detail}
\label{tab:lamp}
\centering
\begin{tabular}{p{2cm} | p{0.7cm} p{8cm} | c}
\toprule
Task                    & \multicolumn{2}{c|}{Input}   & Output            \\
\hline
\multirow{4}{*}{LaMP 2} & ID: & [\textit{id}]                        & \multirow{4}{*}{[\textit{tag}]} \\
                        & Input:  & Which tag does this movie relate to among the following tags? Just answer with the tag name without further explanation. tags: [sci-fi, based on a book, comedy, action, twist ending, dystopia, dark comedy, classic, psychology, fantasy, romance, thought-provoking, social commentary, violence, true story] description: [description]                  &                   \\
                        & Profile: & \{id: [\textit{id}], tag: [\textit{tag}], description: [\textit{description}] \}, $\dots$ &                   \\
\hline
\multirow{4}{*}{LaMP 3} & ID: & [\textit{id}]                         & \multirow{4}{*}{[\textit{score}]} \\
                        & Input & What is the score of the following review on a scale of 1 to 5? just answer with 1, 2, 3, 4, or 5 without further explanation. review: [review],
                     &                   \\
                        & Profile &  \{id: [\textit{id}], tag: [\textit{text}], description: [\textit{score}] \}, $\dots$           \\
\hline
\multirow{4}{*}{LaMP 7} & ID: & [\textit{id}]                             & \multirow{4}{*}{[\textit{tweet}]} \\
                        & Input: & Paraphrase the following tweet without any explanation before or after it: [\textit{tweet}]                 &                   \\
                        & Profile: & \{id: [\textit{id}], tag: [\textit{text}]\}, $\dots$            \\
\bottomrule
\end{tabular}
\end{table*}

\subsection{Prompt Template}
\label{appendix:prompt}

Following is the history and prompt template used to query the base LM to generate preference samples for different LaMP task. History prompt format follows the format used by LaMP benchmark \cite{salemi-etal-2024-lamp}.

\paragraph{LaMP 2: Personalized Movie Tagging} \ 

\textbf{Personalize prompt:} Suppose you are a user with the following user profile history of movie tagging: 
[HISTORY]

Now, given a new description: [QUERY]

Question: Which tag does this movie relate to among the following tags? Just answer with only ONE tag name without further explanation. tags: [sci-fi, based on a book, comedy, action, twist ending, dystopia, dark comedy, classic, psychology, fantasy, romance, thought-provoking, social commentary, violence, true story]

You are a helpfully personalized assistant. You try to predict the movie tagging that the user preferred based on their history. The user prefers [INSIGHT]. Answer only with one tag name (sci-fi, based on a book, comedy, action, twist ending, dystopia, dark comedy, classic, psychology, fantasy, romance, thought-provoking, social commentary, violence, true story). 

Your answer: [OUTPUT] \\
\textbf{Non-personalize/Neutral prompt:} Suppose you are a user with the following user profile history of movie tagging: 
[HISTORY]

Now, given a new description: [QUERY]

Question: Which tag does this movie relate to among the following tags? Just answer with only ONE tag name without further explanation. tags: [sci-fi, based on a book, comedy, action, twist ending, dystopia, dark comedy, classic, psychology, fantasy, romance, thought-provoking, social commentary, violence, true story]

You are a generic and impersonal assistant. You do not consider the user's preferences or profile history when responding. Your answer shoulds [INSIGHT]. Answer only with one tag name (sci-fi, based on a book, comedy, action, twist ending, dystopia, dark comedy, classic, psychology, fantasy, romance, thought-provoking, social commentary, violence, true story). 

Your answer: [OUTPUT] \\
\textbf{History format:}
\begin{enumerate}[topsep=0pt, itemsep=-5pt]
    \item The tag for movie: "\text{[}DESCRIPTION 1\text{]}" is "\text{[}TAG 1\text{]}".
    \item The tag for movie: "\text{[}DESCRIPTION 2\text{]}" is "\text{[}TAG 2\text{]}".
    \item ...
\end{enumerate}

\paragraph{LaMP 3: Personalized Product Rating} \ 

\textbf{Personalize prompt:} Suppose you are a user with the following user profile history of product rating based on the user's review of the product: 
[HISTORY]

Now, given a new review by the user: [QUERY]

Question: What is the rating score of the following review on a scale of 1 to 5? Just answer with 1, 2, 3, 4, or 5 without further explanation. 

You are a helpfully personalized assistant. You try to predict the rating of the product based on the user history ratings. The user prefers [INSIGHT]. Just answer with 1, 2, 3, 4, or 5 without further explanation.

Your answer: [OUTPUT] 

\textbf{Non-personalize/Neutral prompt:} Suppose you are a user with the following user profile history of product rating based on the user's review of the product: 
[HISTORY]

Now, given a new review by the user: [QUERY]

Question: What is the rating score of the following review on a scale of 1 to 5? Just answer with 1, 2, 3, 4, or 5 without further explanation. 

You are a generic and impersonal assistant. You do not consider the user's preferences or profile history when responding. Your answer should [INSIGHT].

Your answer: [OUTPUT] 

\textbf{History format:}

\begin{enumerate} [topsep=0pt, itemsep=-5pt]
    \item \text{[}SCORE 1\text{]} is the rating score for product: "\text{[}TEXT 1\text{]}".
    \item \text{[}SCORE 2\text{]} is the rating score for product: "\text{[}TEXT 2\text{]}".
    \item ...
\end{enumerate}

\paragraph{LaMP 7: Personalized Tweet Paraphrasing} \ 

\textbf{Personalize prompt:} Suppose you are a twitter user with the following user profile history that shows their preferred way of speaking:
[HISTORY]

Now, given a new twitter post: [QUERY]

Question: Paraphrase the tweet in the style the user likes without any explanation before or after it. 

You are a helpfully personalized assistant. You try to paraphrase the tweet in the style the user likes based on the history. The user prefers [INSIGHT].

Your answer: [OUTPUT] 

\textbf{Non-personalize/Neutral prompt:} Suppose you are a twitter user with the following user profile history that shows their preferred way of speaking:
[HISTORY]

Now, given a new twitter post: [QUERY]

Question: Paraphrase the tweet in the style the user likes without any explanation before or after it. 

You are a generic and impersonal assistant. You do not consider the user's preferences or profile history when responding. Your answer should [INSIGHT].

Your answer: [OUTPUT] 

\textbf{History format:}

\begin{enumerate} [topsep=0pt, itemsep=-5pt]
    \item \text{[} TWEET 1 \text{]}
    \item \text{[} TWEET 2 \text{]}
    \item ...
\end{enumerate}

\subsection{Details on Representation Editing}\label{appsubsec:rep_edit}
We provide the details of Section \ref{sec:representation_editing}. We identify personalized and non-personized directions using singular value decomposition (SVD) or contrast consistent search (CCS) on the preference data embeddings. Let $\Phi_l$ represent the function that maps an input sentence to the LM embedding space at layer $l$. For each pair $(p_u^{i, P}, p_u^{i, N})$, we obtain their corresponding representations $\Phi_{l,u}^{i, P}$ and $\Phi_{l,u}^{i, P}$, respectively. To begin, we construct an embedding matrix for personalized direction for user $u$, denoted as $\textbf{H}_{l, u}^{P}$, using these representations:
\[
\textbf{H}_{l, u}^{P} := \left [ \Phi_{l, u}^{1, P} \middle| \ldots \middle| \Phi_{l, u}^{K, P} \right ]^T,
\]
where $K$ is the total number of self-generated data.
Similarly, we create the non-personalized preferences embedding matrix $\textbf{H}_{l, u}^{N}$.

\paragraph{SVD-Based Identification.} Our approach for identifying personalized embedding directions involves using singular value decomposition (SVD) on the preference data embeddings. We extract the top right singular vector of $\textbf{H}_{l, u}^{P}$ as $\theta_{l,u}^P$. Intuitively, we view $\theta$ as the direction that best captures the underlying personalized characteristics. We identify the personalized embedding direction for user $u$ as follows:
\begin{align}
\label{eq:svd}
& \textbf{H}_{l,u}^{P} = \textbf{U}\Sigma\textbf{V} \nonumber \\
& \theta_{l,u}^{P} := \textbf{V}_{0,*}.
\end{align}
Here, $\textbf{U}$ and $\textbf{V}$ represent the left and right unitary matrices produced by running SVD, respectively, and $\Sigma$ is the diagonal matrix of singular values. We define $\theta_{l,u}^{P}$ as the first row of $\textbf{V}$, corresponding to the top right singular vector of $\textbf{H}_{l, u}^{P}$. The non-personalized direction $\theta_{l,u}^{N}$ is defined similarly.

\paragraph{CCS-Based Identification \cite{burnsccs}.} Another approach for identifying these directions is by finding a hyperplane in the latent space that separates personalized data embeddings from non-personalized ones. Typically, this is achieved by training lightweight probes $\theta_{l,u}$ that maps $\Phi_{l,u}^{P}$ and $\Phi_{l,u}^{N}$ to their respective classification labels \cite{li2024inference}. However, we face the challenge of avoiding overfitting to the noise inherent in self-generated data, which limits the applicability of supervised classifier loss in our context. To mitigate this issue, we employ the unsupervised Contrast-Consistent Search (CCS) loss $\mathcal{L}_{CCS}$ proposed in \cite{burnsccs}. Adapting the definition from \cite{burnsccs} to our notations, $\mathcal{L}_{CCS}$ for each user $u$ can be expressed as:

\resizebox{.94\linewidth}{!}{
  \begin{minipage}{\linewidth}
\begin{align*}
\label{eq:ccs}
&\mathcal{L}_{consistency}(g_{\theta, b}, \Phi_{l,u}^{i, P}, \Phi_{l,u}^{i, N})))\\
&:=\left[g_{\theta, b}(\Phi_{l,u}^{N})-(1-g_{\theta, b}(\Phi_{l,u}^{P}))\right]^2 \nonumber \\
&\mathcal{L}_{confidence} (g_{\theta, b}, \Phi_{l,u}^{i, P}, \Phi_{l,u}^{i, N})))\\
&:= \min\left\{g_{\theta, b}(\Phi_{l,u}^{N}), g_{\theta, b}(\Phi_{i,u}^{P})\right\} \nonumber \\
& \mathcal{L}_{CCS}(g_{\theta, b}) := \frac{1}{K}\sum_{i=1}^K ( \mathcal{L}_{consistency}(g_{\theta, b}, \Phi_{l,u}^{i, P}, \Phi_{l,u}^{i, N}) \\
& + \mathcal{L}_{confidence} (g_{\theta, b}, \Phi_{l,u}^{i, P}, \Phi_{l,u}^{i, N})),\\
\end{align*}
\end{minipage}}

where $g_{\theta, b}(v) = \frac{1}{1+e^{-(\theta^\top v+b)}}.$
Training $\theta_{l,u}^N$ with the $L_{CCS}$ objective aims to find a separating hyperplane without fitting any labels with $\mathcal{L}_{consistency}$ and concurrently promoting maximum separation with $\mathcal{L}_{confidence}$.

\paragraph{Hybrid Identification.} While both SVD-based or CCS-based identification can be used to identify both of personalized and non-personalized directions, our exploration revealed that the best results are achieved by combining the two approaches. Specifically, we use SVD to identify $\theta_{l,u}^{P}$ and CCS to determine $\theta_{l,u}^{N}$. This combined approach leverages the strengths of both techniques: SVD's ability to capture the primary direction of personalized embeddings and CCS's effectiveness in identifying the hyperplane that best separates non-personalized embeddings from personalized ones.

\subsection{Time-constrained experiment Set Up}
\label{appendix:time}

\paragraph{$\SYSNAME$} The approximation for the time taken for our experiment is 10, 20, 30 and 40 minutes.

\paragraph{DPO} DPO experiment is trained on $40 \%$, $60 \%$, $80 \%$, $100 \%$ of the LaMP2 partition to get the approximate same time. The  hyperparameters we used consist of 1 training epoch, a batch size of 16, a gradient accumulation step of 1, a learning rate of 5e-5, a max grad norm of 0.3, a warmup ratio of 0.1, a precision of bfloat16, a memory saving quantize flag of "bnb.nf4", a learning rate scheduler type of cosine, and an optimizer of AdamW with PEFT configurations of a r of 256, a $\alpha$ of 128, a dropout of 0.05 and a task type of causal language modeling"

\paragraph{ALOE} We trained ALOE with $7\%$, $23\%$, $39\%$, $55\%$ of the LaMP2 training partition with a relatively equal percentage of CodeAct data \cite{wang2024executablecodeactionselicit} as described by ALOE \cite{wu2024aligning}. We used parameters provided in their SFT hyperparameters, which contains  1 training epoch, a per device train batch size of 1, a gradient accumulation step of 48, a learning rate of 1e-5, and a max sequence length of 8192.

\subsection{Computing Resources}
% \paragraph{Computing Resources} 
All experiments are trained on an Amazon EC2 Instances with eight NVIDIA A100-SXM4-40GB.

\end{document}